\documentclass[letterpaper, 10 pt, conference]{ieeeconf}
\IEEEoverridecommandlockouts
\usepackage{cite}
\usepackage{amsmath,amssymb,amsfonts}
\usepackage{hyperref} 
\usepackage{algorithm}
\usepackage{algpseudocode}
\usepackage{graphicx}
\usepackage{textcomp}
\usepackage{xcolor}
\usepackage{multirow}
\usepackage{booktabs,multirow,makecell}
\usepackage{graphicx} 
\def\BibTeX{{\rm B\kern-.05em{\sc i\kern-.025em b}\kern-.08em
    T\kern-.1667em\lower.7ex\hbox{E}\kern-.125emX}}
\begin{document}

\title{LLM Trainer: Automated Robotic Data Generation via Demonstration Augmentation using LLMs}
\author{Abraham George$^{1}$ and Amir Barati Farimani$^{1}$
\thanks{$^{1}$With the Department of Mechanical Engineering,
        Carnegie Mellon University 
        {\tt\small aigeorge@andrew.cmu.edu, barati@cmu.edu}}%
}

\maketitle

\begin{abstract}
We present \emph{LLM Trainer}, a fully automated pipeline that leverages the world knowledge of Large Language Models (LLMs) to transform a small number of human demonstrations (as few as one) into a large robot dataset for imitation learning. Our approach decomposes demonstration generation into two steps: (1) offline demonstration annotation that extracts keyframes, salient objects, and pose–object relations; and (2) online keypose retargeting that adapts those keyframes to a new scene, given an initial observation. Using these modified keypoints, our system warps the original demonstration to generate a new trajectory, which is then executed, and the resulting demo, if successful, is saved. Because the annotation is reusable across scenes, we use Thompson sampling to optimize the annotation, significantly improving generation success rate. We evaluate our method on a range of tasks, and find that our data annotation method consistently outperforms expert-engineered baselines. We further show an ensemble policy that combines the optimized LLM feedforward plan with a learned feedback imitation learning controller. Finally, we demonstrate hardware feasibility on a Franka Emika Panda robot. For additional materials and demonstration videos, please see the project website: \href{https://sites.google.com/andrew.cmu.edu/llm-trainer}{https://sites.google.com/andrew.cmu.edu/llm-trainer}

\end{abstract}

\section{Introduction}
\begin{figure}[thpb]
      \centering
      \includegraphics[width=\linewidth]{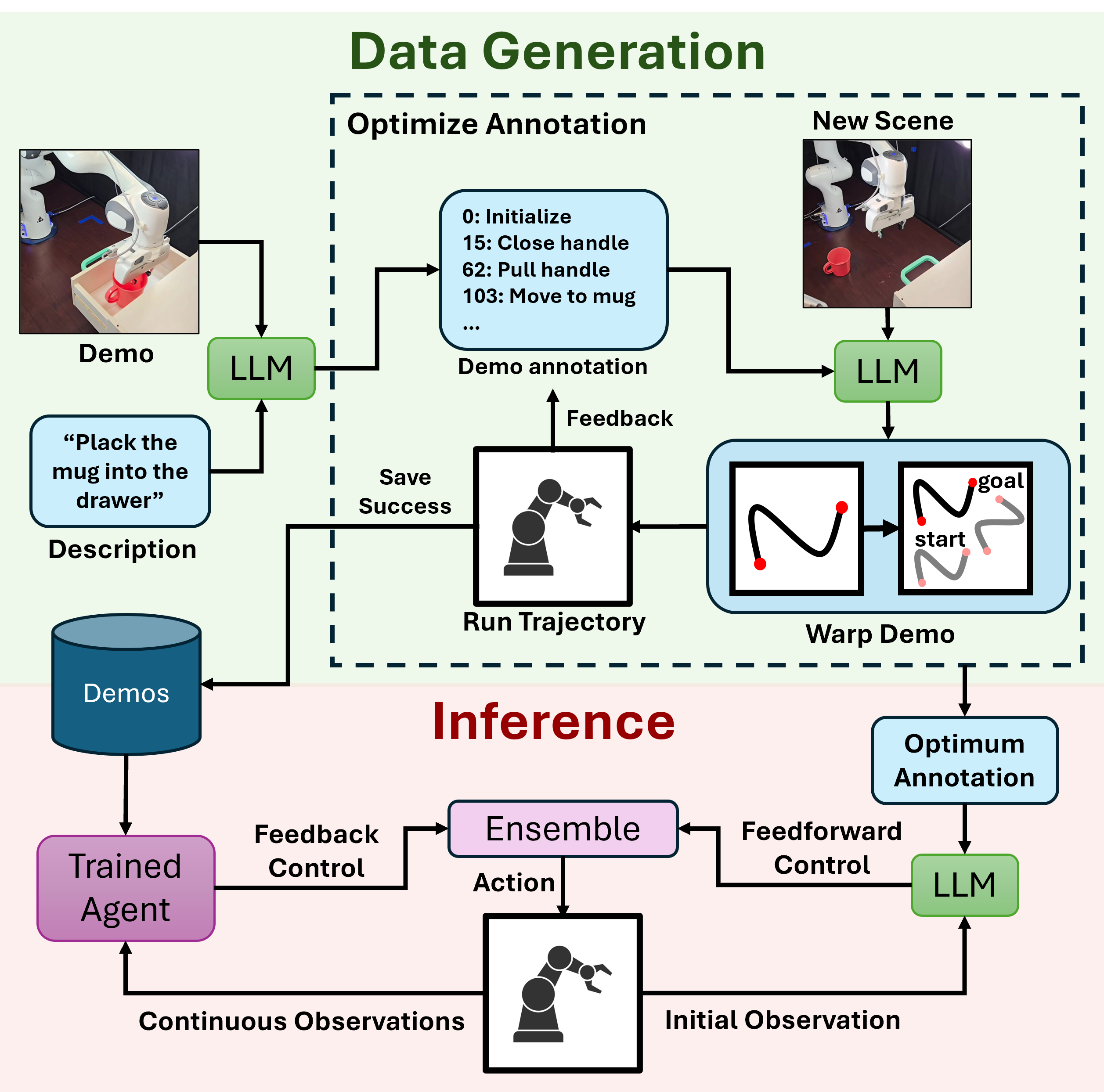}
      \caption{Diagram of our approach. First, the LLM annotates the demonstration, noting key points from the task and any relevant objects. A new instance of the task environment is initialized, and the LLM uses the initial observation and the annotated demo to create a new set of keypoints. These keypoints are used to warp the demonstration trajectory, and the warped trajectory is rolled out in the new environment. If successful, the generated data is saved for later use in training an IL agent. The binary feedback from the rollout (success or failure) is used to optimize the demo annotation.}
      \label{fig:llm-overview}
      \vspace{-0.5cm}
\end{figure}

Recent advances in Large Language Models (LLMs) have revolutionized the field of robot learning, with applications ranging from task planning \cite{saycan2022}, to tool use in long horizon tasks \cite{car2024plato}, to deformable object manipulation \cite{bartsch2024llm}.
At the core of these works is the LLM's broad base of world knowledge, gathered from training on internet-scale data, which allows these agents to be extremely generalizable. In this work, we seek to leverage the world knowledge of LLMs to fully automate demonstration generation through human demo augmentation. To do this, we employ a similar pipeline as \cite{george2023minimizing} and \cite{george2023one} for data generation: first, identify key robot poses in a demonstration, then generate a new environment and modify the key poses based on an initial observation, and finally, use the new key poses to warp the demonstration trajectory, resulting in a new trajectory, which is rolled out in the new environment. However, unlike prior works which rely on human annotation and hard-coded methods to identify key poses and modify them in response to the new environments \cite{george2023minimizing, george2023one, mandlekar2023mimicgen}, our system seeks to fully automate this process by leveraging large language models (LLMs). An outline of our method can be seen in Fig. \ref{fig:llm-overview}. 

Our method for LLM-based data generation has two main steps: First, the LLM annotates the human demonstration, identifying keyframes (timesteps that are important inflection points for the task), listing relevant objects at each keyframe, and explaining the relationship between the robot and these objects. Second, the LLM uses this annotation, along with an initial observation of a newly initialized scene, to determine how the robot's pose should be adjusted at each keypoint. Because the first step of this process does not require information from the new scene, we can reuse these annotations, saving compute cost and opening the door for optimization. By employing a multi-armed bandit-based method, we are able to optimize the demo annotation step, improving data generation success rate by 2-3 times. 

Once the data generation process is complete, we can use the generated data to train imitation learning agents. Additionally, thanks to our annotation optimization process, we develop a highly effective LLM-based feedforward policy during data generation. In addition to serving as a viable agent on its own, we show that this feedforward policy, when combined with the feedback agent, can form an effective ensembled policy, combining the long-horizon planning and generalizability of LLMs with the feedback control of imitation learning. 

This work has three main contributions:
\begin{itemize}
    \item An LLM-based data generation method that autonomously generates data using only a single, unannotated demonstration and a short (one sentence) description of the task. 
    \item A multi-armed bandit based optimization method which significantly improves demo generation success rate, allowing our method to outperform baselines that rely on expert annotations.
    \item An ensembling strategy to combine a learned IL policy with the optimized LLM-based feedforward controller developed during data collection.
\end{itemize}

\textbf{Terminology Note:} The models used in this work accept both text and image inputs. While such models are sometimes referred to as Vision Language Models (VLMs), we use the term LLM for consistency with prior literature.

\section{Related Works}

\subsection{Automated Robotic Demonstration Generation}
Demonstration generation has emerged as a critical tool for addressing the large data requirements of robot learning. \cite{george2023minimizing} first proposed using manually labeled anchor points tied to object locations to programmatically warp recorded trajectories to generate novel demonstrations, and used this method to assist in training an RL agent, both through replay buffer spiking \cite{lipton2016} and through curriculum learning \cite{10.1145/1553374.1553380}. MimicGen \cite{mandlekar2023mimicgen} employed a similar approach to programmatically adapt a set of recorded demonstrations, using human annotations to separate each demonstration into object-specific sub-trajectories, then warp each sub-trajectory based on the change in the relevant object's pose. MimicGen evaluated their method using evaluation tasks and imitation learning benchmarks from RoboMimic \cite{robomimic2021}. Similar to MimicGen, OneACTPlay also used data generation to train imitation learning agents using only a single human demonstration \cite{george2023one}. Additionally, this work developed a novel uncertainty-based addition to Action Chunking Transformer's (ACT) \cite{zhao2023learning} temporal ensembling method to better handle out-of-distribution states. Additional follow-up works to MimicGen have extended this methodology to bimanual manipulation \cite{jiang2025dexmimicgen} and have incorporated hybrid control, using motion planners to assist in the generation of new trajectories \cite{garrett2024skillmimicgen}. However, as our work does not explore bimanual manipulation, and we seek to avoid human annotation, hard-coding, and embodiment-specific methods, we use \cite{george2023one} and \cite{mandlekar2023mimicgen} as our baselines. 

\subsection{LLMs for Robot Control and Data Generation}
Recent advances in Large Language Models (LLMs) have enabled transformative progress in many aspects of robot control \cite{li2025large, kim2024survey, wang2025large}. Early work, such as SayCan \cite{saycan2022}, combined LLM-based reasoning with robot affordance grounding to connect abstract language instructions to predefined robot skills. More recent works have extended LLM-based control to a wide range of methodologies, ranging from code-based control \cite{liang2022code} to path and task planning \cite{merrill2025llm, barkley2025semantic}. In addition to direct robot control, LLMs have been used to assist in policy training, with a key avenue being through improved data generation. \cite{katara2024gen2sim} used LLMs to automate the generation of 3D assets, task descriptions, and reward functions in simulation, creating novel tasks to train RL agents. Similarly, \cite{wang2023gensim} used LLMs to iteratively generate novel simulated tasks, modifying an existing environment either to adapt it to meet a desired task description or to explore novel tasks. \cite{hua2024gensim2} extended this work to handle long-horizon tasks with articulated objects. Likewise, \cite{jing2025humanoidgen} used LLMs to automate task generation for humanoid robots. As with \cite{katara2024gen2sim}, these works utilize their generated environments to train robotic agents, using reinforcement learning, pre-trained solvers, motion planning, or LLM-generated expert code. However, none of these methods explicitly leverage human demonstrations or the keypoint-based demonstration augmentation methods used by MimicGen (and its related family of works) and OneACTPlay. In our paper, we seek to close this gap by combining LLMs with demo augmentation-based data generation.

\section{Methods}
\subsection{Key Assumptions}
\label{sec:key_assumptions}
As with \cite{mandlekar2023mimicgen, george2023one}, we assume that we have a set of demonstrations, $\mathcal{D} = \{D_1, D_2, \dots, D_N\}$, each consisting of a set of observations $O = \{o_t\}_{t=1}^T$, robot poses $P = \{p_t\}_{t=1}^T$, and actions $A = \{a_t\}_{t=1}^T$. Additionally, we assume that the actions are in the form of goal end-effector poses and the observations consist of image observations of the scene, the pose of key objects in the scene, and the robot's pose. We also assume that we have access to an initial observation from each new scene, $o_1^{\text{new}}$, including both image observations and object poses. Because this observation only applies to the initial observation of the new scene, we can extract this observation data using annotation tools that would not be practical during operation (i.e., LLM Description + SAM \cite{kirillov2023segment} + Grounding Dino \cite{liu2023grounding}). Finally, we assume a short (one or two sentence) description of the task, $s$.

\subsection{Identifying and Modifying Key Poses with LLMs}
We can view key point identification and modification as a functional mapping of a demonstration, task description, and new observation to a modified set of key points $\mathcal{F} : (D, s, o_1^{\text{new}}) \mapsto K'$.

We can decompose this function into two subfunctions, one which extracts keypoints and a separate function that modifies these keypoints $\mathcal{F} : (D, s) \mapsto K; \quad \mathcal{G} : (D, s, o_1^{\text{new}}, K) \mapsto K'$. Since $\mathcal{F}$ does not require information on the new scene, we only need to run $\mathcal{F}$ once during data generation. We can take further advantage of this by having $\mathcal{F}$ return processed demonstration information $\hat{D}$, which can replace the original demonstration $D$, thereby shifting this processing from $\mathcal{O}(T)$ to $\mathcal{O}(1)$. Combining these two modifications, we define the following composite functions: $\mathcal{F} : (D, s) \mapsto K, \hat{D}; \quad \mathcal{G} : (\hat{D}, s, o_1^{\text{new}}, K) \mapsto K'$.

\begin{figure}[thpb]
      \centering
      \includegraphics[width=\linewidth]{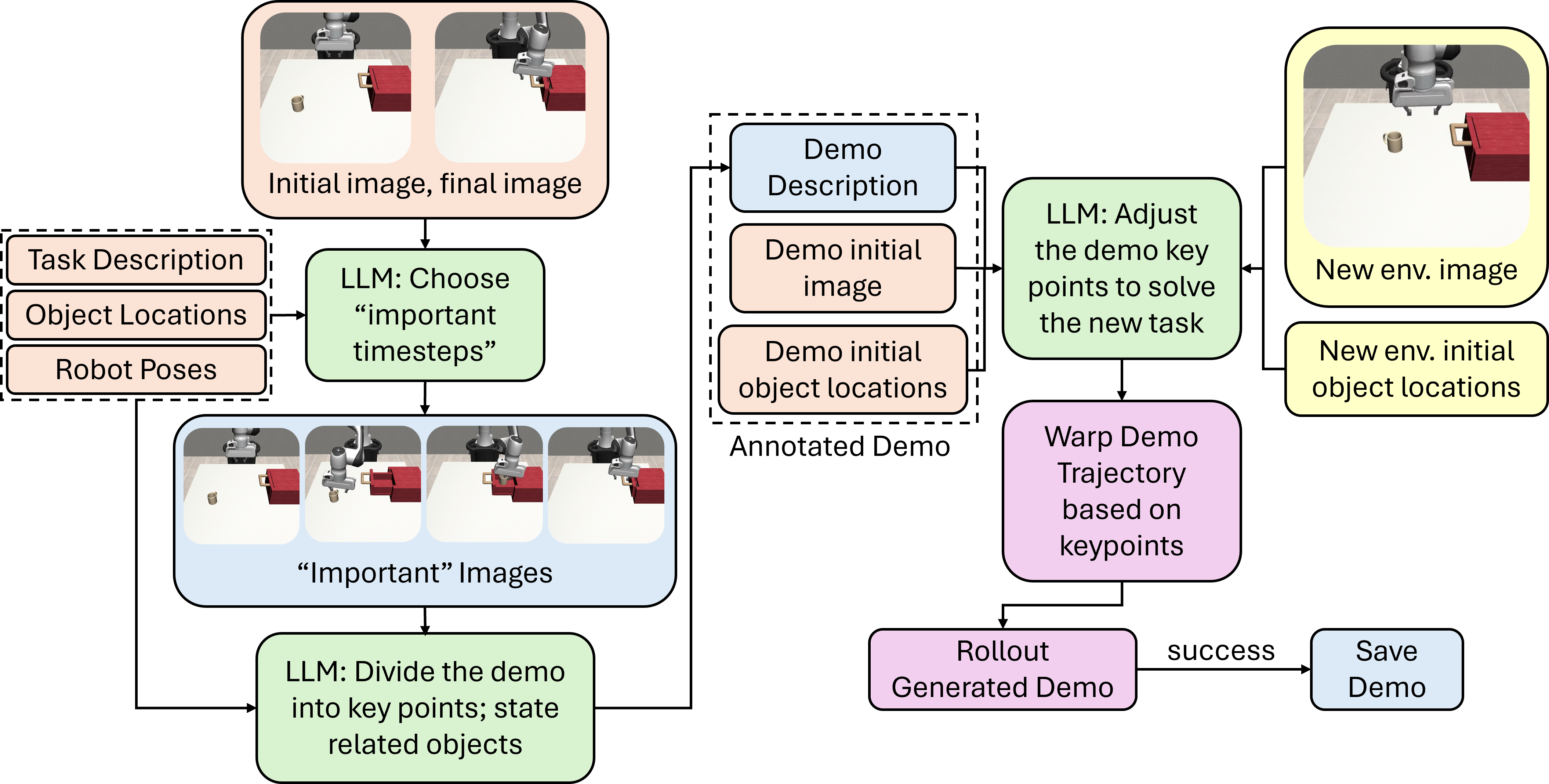}
      \caption{Diagram of the LLM-based demo augmentation method. First, the LLM is used to annotate the demonstration, noting key points and the relationship between these points and objects in the scene. Next, the LLM uses this annotation to adapt the key points to a new scene, and then uses the new key points to warp the recorded trajectory. The recorded trajectory is rolled out, and if successful, saved. }
      \label{fig:llm_pipeline}
\end{figure}

We leverage the world knowledge of LLMs for keypoint identification and mapping, using LLMs to perform $\mathcal{F}$ and $\mathcal{G}$. A detailed diagram of our approach can be seen in Fig. \ref{fig:llm_pipeline}. To perform these functions using an LLM, we must convert the demonstration into a compatible form (images and text) and convert the LLMs output (text) to keypoints. Although demo conversion is straightforward (the visual observations can be presented as images, and the robot and object poses for each timestep can be converted to text), the resulting LLM input would be far too large (especially the images), as each demo has hundreds of timesteps. Instead, we can have the LLM choose which timesteps are worth viewing. We provide the LLM with the initial and final observations from a single scene overview camera and the pose of the robot and the objects throughout the demo. 

To reduce the amount of text the LLM has to parse, we only provide the object poses every 5 timesteps (approximately).  We add a small amount of noise to this sampling process, so the list of timesteps is slightly different for each run to keep the LLM from attaching to specific timestep numbers. All poses are presented to the LLM as a translation (mm) and a rotation expressed in Euler angles (deg). Additionally, the end-effector's rotation is presented as the relative rotation (in the world coordinate frame) from the end-effector's home pose, which we define as the first pose of the first demonstration in $\mathcal{D}$.
\begin{equation}
\mathbf{R}_{\text{LLM},t} = \mathbf{R}_{\text{home}}^{-1} \mathbf{R}_{t}
\end{equation}
We found that the LLM struggles with reasoning about rotation, and using relative rotation mitigates the impact of unintuitive default rotations.

Given this demonstration summary and the task description, the LLM responds with a list of timesteps it views as important. Images and object poses for these timesteps, along with the full robot trajectory and the summary for the task, are passed to the LLM (a new instance, separate from the prior query), which returns a description of the demonstration, including a list of key poses (both the timestep and the pose), a list of important objects for each key pose, and instructions on how the key poses should be modified in response to alterations in these objects' poses. Occasionally, the LLM will not have perfect recall and incorrectly list the robot's pose at a key point. In these cases, we replace the listed key pose with the recorded robot pose at the specified timestep (both in the extracted list of key poses and the text of the demo description). Additionally, we add the initial and final poses of the demonstration as key poses if the LLM did not list these points. If the LLM includes any poses with a timestep outside of the range of the demonstration, fails to match the response formatting criteria, or some other error occurs during processing, we discard the LLM responses and restart the annotation generation process. The final key point list, $K$, and processed demo description text are combined with the original demonstration information to form an \textit{annotated demo} and saved.

Next, we use the annotated demo to perform $\mathcal{G}$ using an LLM. First, we initialize a new environment and capture the initial observation $o_1^{\text{new}}$. From this observation, we extract an image of the new scene and the initial poses of the robot and the objects in the scene. We then combine the initial robot pose, initial object poses, initial image, final image, and demo description text (which includes the key points) from the demo annotation to form $\hat{D}$. $\hat{D}$, the task description, $s$, and the processed initial observation, $o_1^{\text{new}}$, are combined and passed into the LLM, which responds with a list of new keypoints, $K'$, adjusted based on the new observation in accordance with the instruction included in $\hat{D}$. These new keypoints are then used to warp the demonstration trajectory, and the warped trajectory is executed in the new environment to generate a novel demonstration.

\subsection{Key-Pose Based Trajectory Warping}
We use the same warping mechanism proposed in \cite{george2023one}, with a minor addition to handle end-effector rotation. Given a demonstration trajectory and a new start and end position, $\mathbf{p}^{\mathrm{new}}_0$ and $\mathbf{p}^{\mathrm{new}}_T$, we generate the rigid transform, $\mathbf{T}_{\text{warp}}$, which, when applied to the demonstration trajectory, aligns the demonstration start and endpoints, $\mathbf{p}^{\mathrm{old}}_0$ and $\mathbf{p}^{\mathrm{old}}_T$, with the new start and end positions. This transform has an extra degree of freedom (rotation about the axis from the start point to the end point), which we use to ensure that the transformed z-axis is aligned with the world's z-axis, $\hat{z}$. 
\begin{equation}
\mathbf{T}_{\mathrm{warp}} =
\begin{bmatrix}
\mathbf{R}_{\mathrm{warp}} & \mathbf{t} \\
\mathbf{0}^\top & 1
\end{bmatrix},
\quad \text{s.t. }
\begin{cases}
\mathbf{p}^{\mathrm{new}}_0 = \mathbf{T}_{\mathrm{warp}}\,\mathbf{p}^{\mathrm{old}}_0 \\[4pt]
\mathbf{p}^{\mathrm{new}}_T = \mathbf{T}_{\mathrm{warp}}\,\mathbf{p}^{\mathrm{old}}_T \\[6pt]
\max \; \hat{\mathbf{z}}^\top \mathbf{R}_{\mathrm{warp}} \hat{\mathbf{z}}
\end{cases}
\end{equation}
where $\mathbf{p}$ is a position in homogeneous coordinates. 

This transformation is then applied to the demonstration trajectory to form a new trajectory: 
\begin{equation}
    \mathbf{p}^{\mathrm{new}}_t = \mathbf{T}_{\mathrm{warp}}\,\mathbf{p}^{\mathrm{old}}_t
\end{equation}

When working with robot poses (position + rotation), we apply a similar methodology to warp the robot's positions. For rotations, we calculate the delta rotation between the demonstration trajectory and the new trajectory for the start and end points:
\begin{equation}
\Delta \mathbf{R}_{0} = \mathbf{R}^\text{new}_0 (\mathbf{R}^\text{demo}_0)^{-1}
\quad
\Delta \mathbf{R}_{T} = \mathbf{R}^\text{new}_T (\mathbf{R}^\text{demo}_T)^{-1}
\end{equation}
then linearly interpolate between them to form a delta rotation for each timestep in the trajectory segment $\Delta \mathbf{R}_t$. Finally, these delta rotations are applied to the demonstration rotation to form the warped rotation:
\begin{equation}
\mathbf{R}_t^{\text{new}} = \Delta \mathbf{R}_t \mathbf{R}_t^{\text{demo}}
\end{equation}

\subsection{Data Collection via Multi-Armed Bandit Optimization}
To create a dataset to train a robotic agent, we repeat the above demo augmentation method many times, rolling out each generated trajectory and saving successful runs as novel demonstrations. As discussed above, the demo generation method consists of two steps: first we annotate the given demonstration, then we generate a new demonstration using this annotation and the initial observation from the new environment. Because this annotation includes both the key poses to be warped and the instructions for how to warp them, the success of the second step relies heavily on a high-quality annotation. As such, if we wish to optimize the data generation process, we must generate, identify, and use a high-quality annotation. With this in mind, we cast the data generation process as a modified multi-armed bandit problem. Here, each ``arm" is a demo annotation, and our goal is to determine which annotation we should use to generate a demo to maximize the total number of successful demos, balancing exploration (experimenting to determine which annotation is best) with exploitation (using the best annotation). In this process, we are optimizing the success rate of our generated demo (ie. minimizing the number of rollouts required to generate a set number of successful demonstrations). This ignores the cost of running the LLM. We believe that this is the right choice for our analysis, as our emphasis is on robot demo generation, reducing the number of robot rollouts will (generally) reduce the amount of LLM compute, and equating LLM compute to robot rollouts in a general case is infeasible. 

Because this multi-armed bandit problem has a binary reward (each rollout either succeeds or fails), we can employ Thompson Sampling \cite{russo2018tutorial}. This method models each arm as a beta distribution, where the success rate of each arm is treated as a Bernoulli process. We assume that the prior distribution is uniform, resulting in a beta distribution for each arm of the form:
\begin{equation}
P(\text{arm}_i) \sim \text{Beta}(n_{suc, i} + 1,\; n_{fail, i} + 1)
\end{equation}
where $n_{suc, i}$ is the number of successes and $n_{fail, i}$ is the number of failures recorded for arm i.

At each iteration, we sample a success probability from each annotation's beta distribution and select the annotation with the highest sampled value to generate the next demonstration. This approach naturally balances exploration and exploitation: underexplored annotations with high uncertainty have a greater chance of being selected early on, while consistently successful annotations are chosen more frequently overall. As more demonstrations are collected, the probability estimates converge, leading to more reliable selections and an improved overall success rate in data generation.

However, our problem differs from the standard multi-armed bandit formulation in that, at each step, we have an additional option - we can attempt to generate a new demo annotation. To do this, we sample a random demonstration from $\mathcal{D}$, and use it to generate a new demo annotation. We then immediately generate a new trajectory using the annotation, roll it out, and see if it is a success. If the new annotation results in a success, we add it to the bandit. If not, we discard it. To determine whether we should use an existing annotation or generate a new one, we must determine the expected value of each operation, keeping in mind the effect of including a new arm on all future Thompson Sampling steps. If we are going to sample the bandit T more times, the expected value of adding a new annotation is:
\begin{equation}
\mathbb{E}_{add} = P_{\text{add}} \cdot (1 + \mathbb{E}_{T\text{-}1} (\mathcal{P} \cup \{p_{\text{new}}\})) + (1 \text{-} P_{\text{add}}) \cdot \mathbb{E}_{T\text{-}1} (\mathcal{P})
\end{equation}
where $P_{\text{add}}$ is the probability of successfully generating a new annotation (getting a success on the first generation using the new annotation), which we estimate from previous annotation generation attempts, and $\mathbb{E}_{T}(\mathcal{P})$ is the expected value of running Thompson sampling for $T$ iterations using annotations with success probabilities $\mathcal{P} = \{p_1, p_2, \dots, p_n\}$.

Therefore, we should try sampling a new annotation if and only if $\mathbb{E}_{add} > \mathbb{E}_{T}(\mathcal{P})$. To evaluate this, we must find a suitable method to estimate the new annotation's success rate $p_{new}$, and determine the estimated value of a Thompson Sampling rollout. We model the distribution of annotation success rates as a beta distribution. However, directly fitting a beta distribution to the observed arm success rates is challenging, as we only have a small number of annotations (as few as two). Instead, we leverage the fact that we have an estimated success rate distribution for each annotation, and fit the new arm prior to these existing success rate distributions. Formally, this means finding the beta distribution parameters $\hat{\alpha}$ and $\hat{\beta}$ that satisfy:
\begin{equation}
\label{eq:arm_prior}
(\hat{\alpha}, \hat{\beta}) = \arg\max_{\alpha, \beta} \prod_{i=1}^{n} \mathbb{E}_{x \sim \mathrm{Beta}(\alpha_i, \beta_i)} \left[ \frac{x^{\alpha \text{-} 1}(1 \text{-} x)^{\beta \text{-} 1}}{B(\alpha, \beta)} \right]
\end{equation}
where $\alpha_i$ and $\beta_i$ are the beta distribution parameters for arm $i$.

Practically, we evaluate this by sampling a large number ($m$) of success rates from each of the $n$ annotations, then fitting a Beta distribution to the resulting $n \cdot m$ samples.

To evaluate the estimated value of a Thompson sampling rollout $\mathbb{E}_{T}(\mathcal{P})$, we use a brute-force large-number approximation. We begin by sampling $k$ sets of discrete probabilities from the annotation success rate distributions (the beta distributions for each annotation used for Thompson Sampling). For each of these $k$ sets, we run a virtual Thompson Sampling rollout for $T$ iterations, using the $k$ sampled probability sets as ground-truth success rates. If we are evaluating a newly generated annotation, then the success probability is sampled from the prior distribution (equation \ref{eq:arm_prior}), and the annotation is initialized with 1 success and 0 failures (since for a newly generated annotation to be used, its first pull must be successful). When determining if a new annotation should be added, we need to run three of these rollouts: $\mathbb{E}_{T}(\mathcal{P})$, $\mathbb{E}_{T-1}(\mathcal{P})$, and $\mathbb{E}_{T-1}(\mathcal{P}\cup \{p_{\text{new}}\})$. To reduce the impact of the sampling method, we use the same $k$ samples of $\mathcal{P}$ for all three calculations and choose a large value for $k$ (1000). This method requires an estimate of the total number of Thompson Sampling steps that will be performed, $T$. However, we typically run the data generation system until we reach a total number of successful demonstrations. In this case, we approximate $T$ as $T \approx (n_{\text{suc, goal}} - n_{\text{suc, cur}})/\text{max}(\mathcal{P})$.

\subsection{Ensembling}
\begin{figure}[thpb]
      \centering
      \includegraphics[width=\linewidth]{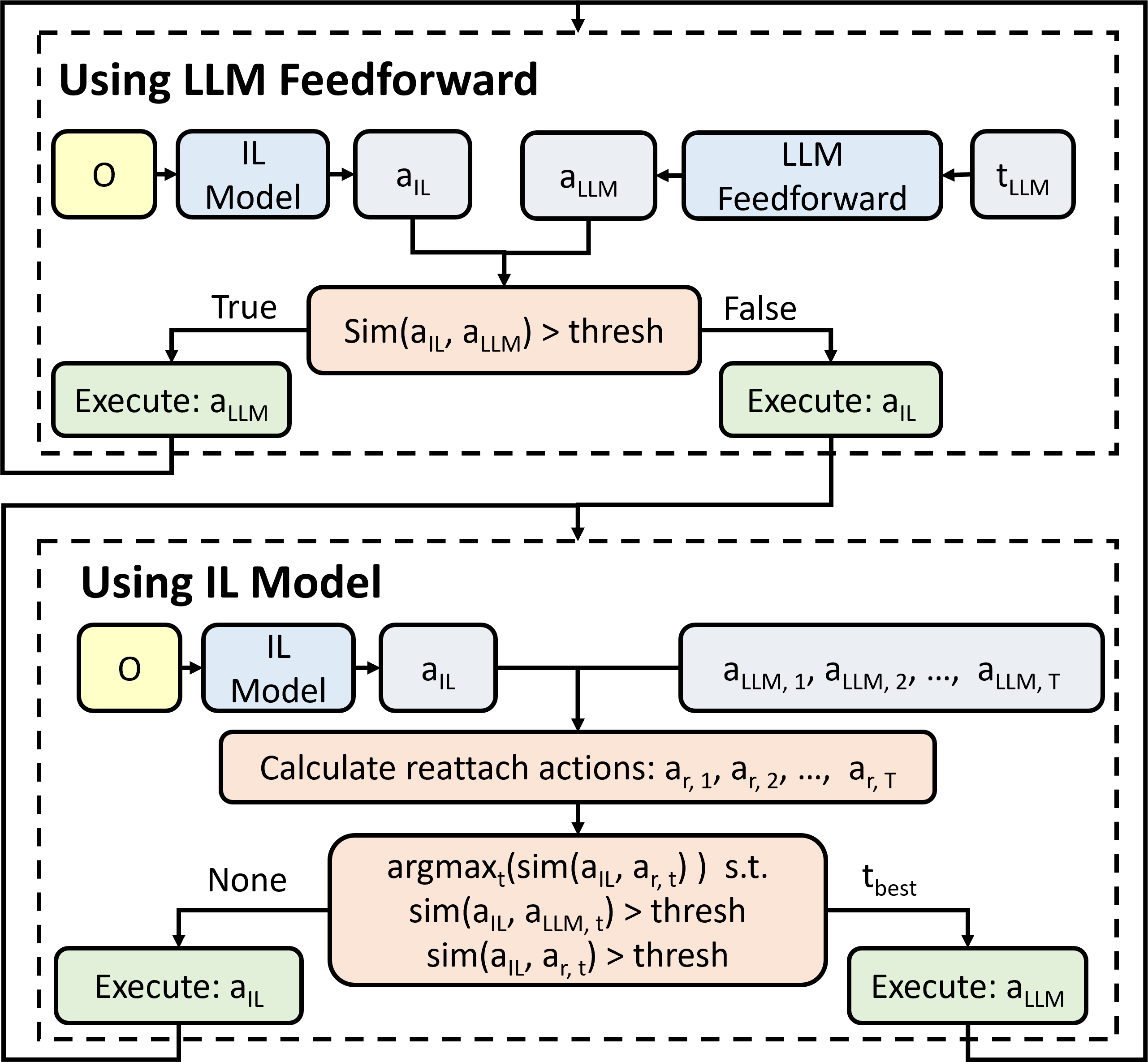}
      \caption{Outline of the ensembling method. The agent begins by running the LLM-based feedforward policy. When the agent has made an error (as determined by increased disagreement between the two policies), the IL agent takes control. Once the agent returns to the feedforward path, the feedforward agent retakes control.}
      \label{fig:ensembling}
\end{figure}

Due to the multi-armed bandit optimization method, our LLM feedforward control scheme serves as a capable agent, reaching success rates on par with and even surpassing those of trained IL agents. Despite their strong performance, these policies have a fatal flaw: as a pure feedforward method, they cannot respond to disturbances in the environment. The trained IL agent solves this problem, but struggles with longer horizon tasks and generalizing to out-of-distribution observations, two tasks that the LLM-based feedforward policy excels at. We take advantage of the complementary strengths of these two policies by ensembling them together. An outline of our ensembling method can be seen in Fig. \ref{fig:ensembling}. Since the feedforward policy excels at high-level planning, the ensembled agent begins by executing the LLM feedforward trajectory. This continues until the ensembled agent detects that the feedforward policy has made an error, at which point the agent switches to the IL policy to correct the error. Once the error is fixed, the ensemble policy can ``reattach" to the feedforward trajectory at an appropriate point and continue executing the pre-planned trajectory until another error is detected. A simple heuristic for determining when the feedforward policy has made an error is the similarity of the action predicted by the feedforward and feedback policies. If the two policies are in substantial prolonged disagreement, then it stands to reason that an error has occurred and the feedback policy should be used. To measure similarity, we use a magnitude-aware cosine similarity between normalized actions (normalization allows us to combine different action modalities). 
\begin{equation}
    \text{sim} =
    \underbrace{\frac{2\min(\|\mathbf{a}_{LLM}\|_{1},\|\mathbf{a}_{IL}\|_{1})}
    {\|\mathbf{a}_{LLM}\|_{1}+\|\mathbf{a}_{IL}\|_{1}}}_{\text{magnitude agreement}}
    \;\cdot\;
    \underbrace{\frac{\mathbf{a}_{LLM}\cdot\mathbf{a}_{IL}}
    {\|\mathbf{a}_{LLM}\|_{2}\,\|\mathbf{a}_{IL}\|_{2}}}_{\text{cosine similarity}}
\end{equation}

After the IL policy recovers from an error, the agent attempts to rejoin the feedforward trajectory. To do this, it evaluates each future point on the recorded trajectory as a potential reattachment target. For each point, we compute (i) a reattach action (the action needed to move from the agent’s current pose to the candidate goal pose) and (ii) the recorded action that the feedforward agent would take at that timestep (based on the recorded trajectory). We then select the first point that satisfies two conditions:
(1) the reattach action is similar to the IL policy’s current action, and
(2) the recorded action is also similar to the IL policy’s current action.
Among all such points, we choose the one with the highest reattach action similarity. To prevent rapid switching, we enforce a short cooldown (5 timesteps) after each policy change before another can occur.
\begin{equation}
    t^* = \arg\!\max_{t}
    \;\text{sim}(\mathbf{a}^{\text{att}}_t,\mathbf{a}^{\text{IL}})
    \;\text{s.t.}\!
    \begin{cases}
        \text{sim}(\mathbf{a}^{\text{att}}_t,\mathbf{a}^{\text{IL}}) > \tau \\
        \text{sim}(\mathbf{a}^{\text{rec}}_t,\mathbf{a}^{\text{IL}}) > \tau
    \end{cases}
\end{equation}
where $t^*$ is the reattach timestep, $\mathbf{a}^{\text{IL}}$ is the action predicted by the imitation learning agent, $\mathbf{a}^{\text{att}}_t$ is the reattach action, $\mathbf{a}^{\text{rec}}_t$ is the recorded action, and $\tau$ is a similarity threshold.

\subsection{Hardware Implementation}
As described in section \ref{sec:key_assumptions}, our system assumes that the observations, $O$, in our set of human demonstrations $\mathcal{D}$, along with the initial observation, $o^{new}_1$, of the new scene, contain the poses of the objects in the environment. Although these observations are easy to access in simulation, if we wish to extend our method to hardware, we need some way to calculate these poses. Furthermore, since the goal of this work is to develop a task-agnostic method for data generation that does not require manual human annotation (aside from a brief textual description of the task), our method for gathering $O$ must also be task-agnostic and fully automated. To achieve this, we implement a system similar to one used in \cite{car2024plato}, using a combination of LLM descriptions, SAM \cite{kirillov2023segment}, and Grounding DINO \cite{liu2023grounding} to extract object locations from RGB-D (color + depth) images of the scene. 

 We assume that a portion of the image observations from each timestep, $\{i^d_t\} \in o_t$, are RGB-D images (color + depth), and that these RGB-D images are sufficient to observe the full scene. As with the simulation experiments, we also assume one of the images provides an overview of the full scene, $i^o_t \in o_t$ (this image does not have to contain depth information). First, we use the LLM to identify the relevant objects for the task, requesting both the name and color of the desired objects. With this list of relevant objects, we can use SAM \cite{kirillov2023segment} and Grounding DINO \cite{liu2023grounding} to extract a segmentation mask of each object from the RGB-D images, which we then use to create a point cloud of the objects.

For our method, we need to calculate the pose of each object for each observation in the human demonstrations and for the first observation of each new scene. To do this, we first define a reference point cloud for each object. Any observation from a human demo, $o_t$, can be used for this, so long as the relevant object is visible (detected by Grounding DINO). For our implementation, we default to the first observation in the demo where the object is visible. We define the pose of the reference point cloud, setting its rotation to zero and its position to the center of the point cloud. To determine the pose of an object in a new observation, we generate its point cloud, then use RANSAC \cite{fischler1981random} and ICP \cite{besl1992method} to determine the transformation from the reference point cloud to the observed point cloud, and apply this transformation to the reference point-cloud's pose. This method helps to ensure consistency in pose measurements across observations.
\begin{equation}
    P_{\text{obs}} = T^{\text{ref}}_{\text{obs}} \; P_{\text{ref}}, \qquad P_{\text{ref}} =
\begin{bmatrix}
\mathbf{I}_{3\times 3} & \mathbf{c}_{ref} \\
\mathbf{0}_{1\times 3} & 1
\end{bmatrix}
\end{equation}

\begin{figure}[thpb]
      \centering
      \includegraphics[width=\linewidth]{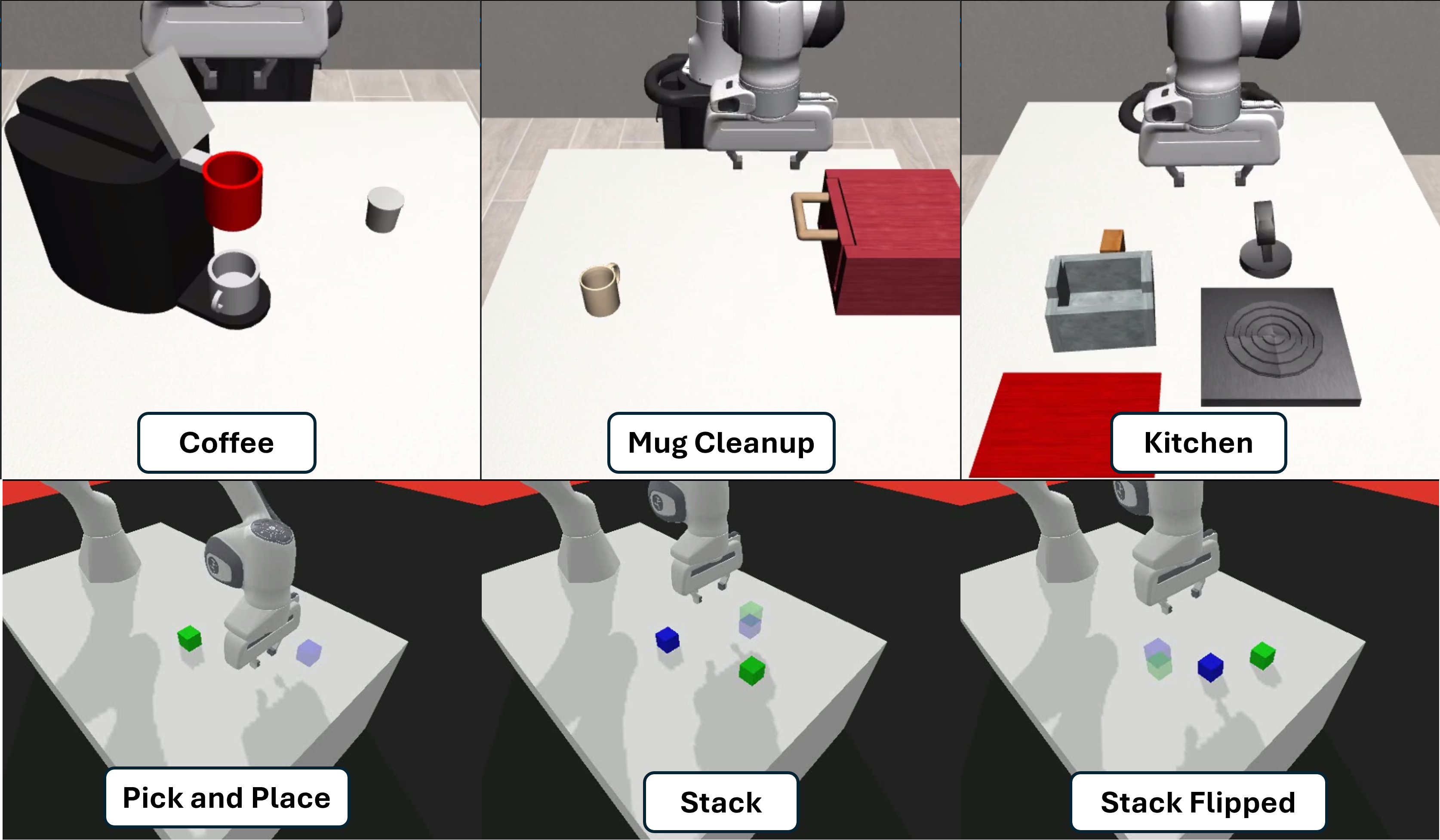}
      \caption{Simulation evaluation tasks. Top: RoboMimic tasks from the MimicGen benchmark. Bottom: PandaGym tasks from the OneACTPlay benchmark}
      \label{fig:sim-envs}
\end{figure}

\section{Results}
To evaluate our proposed method, we use three of the RoboMimic \cite{robomimic2021} simulation environments from MimicGen \cite{mandlekar2023mimicgen}, and the block pick-and-place and block stacking Panda-Gym \cite{gallouedec2021} tasks from OneACTPlay (we did not use the push task as it is a simplified case of the pick-and-place task). Additionally, we included two variations of the block stacking task: block stack flipped and block stack walking. In the ``flipped" version of the stack task, the order in which the blocks must be stacked is flipped for 50\% of runs. The only way this change is conveyed to the agent is the switching of the colors of the transparent goal location blocks in the agent's visual observation - there is no explicit text or flag informing the agent of the change. This task was included to illustrate the reasoning capabilities of the LLM. In the walking version of the stack task, the blocks undergo a random walk, moving 0.4 mm in a random direction each timestep while on the ground. Once they are picked up by the robot, they stop moving. This task was included to see how well the trained agents handled time-varying tasks, and was used to evaluate agents trained on the regular stack task. A diagram showing the evaluation tasks can be seen in Fig. \ref{fig:sim-envs}. For these tasks, we used the same human demonstrations used in \cite{mandlekar2023mimicgen} and \cite{george2023one}. Therefore, for the RoboMimic tasks, we used 10 human demonstrations, and for the OneACTPlay block manipulation tasks, we used a single human demonstration. Throughout our experiments, we use GPT-4o (OpenAI) as the LLM.

\begin{table*}[t]
  \caption{Task descriptions used in our experiments.}
  \label{tab:task-descriptions}
  \centering
  \small
  \setlength{\tabcolsep}{6pt}
  \renewcommand{\arraystretch}{1.15}
  \resizebox{\textwidth}{!}{%
  \begin{tabular}{l p{0.78\textwidth}}
    \toprule
    \textbf{Task} & \textbf{Description} \\
    \midrule
    Mug Cleanup & Open the drawer, pick up the mug, place it in the drawer, then close the drawer. \\
    Kitchen & Turn on the stove, place the pot on the stove, add the bread to the pot, move the pot from the stove to the serving area, then turn off the stove. \\
    Coffee & Pick up the coffee pod, place it into the coffee machine, and close the coffee machine lid.
    \\
    Stack & Place the two blocks in the highlighted region, stacking the green block on top of the blue block. \\
    Stack Color Flip & Place the two blocks in the highlighted region, stacking them on top of each other. 
    \\
    Pick \& Place & Pick up the green block and place it in the blue highlighted region. \\
    \bottomrule
  \end{tabular}}
\end{table*}

Using our proposed method, we generate 1000 successful demos for each MimicGen task, and 400 successful demonstrations for each OneACTPlay task (these values were chosen to match the baselines). We report the generation success rate when using the best annotation found during the multi-armed bandit exploration, the average generation success rate when using a new annotation (this would be the success rate of our method if we did not optimize the annotation), and the overall success rate for the full data generation process, which reflects both the exploration and exploitation steps of the optimization process. Additionally, we used these generated demonstrations to train imitation learning agents and reported the trained agent's performance on each task for a variety of different numbers of demonstrations. For evaluation, we calculate the success rate using 50 evaluation rollouts and report the mean and standard deviation across 5 trials (the policy is trained separately for each trial). Our results, and the comparison to baseline, can be seen in Table \ref{table:gen_suc_rates}. For the MimicGen-trained agent success rates, we follow the advice from their repository and retrain and re-evaluate the agents using their data, rather than use the success rate from the paper, due to discrepancies in the simulator. 

\begin{table}[t]
\centering
\caption{Data generation success rates (\%) on MimicGen and OneACTPlay tasks.}
\label{table:gen_suc_rates}
\small
\setlength{\tabcolsep}{6pt}
\renewcommand{\arraystretch}{1.2}
\begin{tabular}{lcccc}
\toprule
Task & \makecell{No RL} & \makecell{Best\\Annotation} & Total & \makecell{Baseline} \\
\midrule
Mug Cleanup & 12\% & 44\% & 36.1\% & 29.5\% \\
Kitchen & 20\% & 60\% & 50.5\% & 42.7\% \\
Coffee & 39\% & 89\% & 82.6\% & 78.2\% \\
\midrule
Pick \& Place & 65\% & 92\% & 84.5\% & 81.7\% \\
Stack & 34\% & 89\% & 79.1\% & 80.0\% \\
Stack Flipped & 14\% & 58\% & 40.9\% & N/A \\
\bottomrule
\end{tabular}
\end{table}

\begin{table*}[t]
  \caption{Task success rates (\%). Values are mean $\pm$ standard deviation}
  \label{tab:all-demo-results}
  \centering
  \small
  \setlength{\tabcolsep}{4pt}
  \renewcommand{\arraystretch}{1.1}

  \resizebox{\textwidth}{!}{%
  \begin{tabular}{l *{5}{ccc}}
    \toprule
    \multirow{2}{*}{Task} &
      \multicolumn{3}{c}{50 demos} &
      \multicolumn{3}{c}{100 demos} &
      \multicolumn{3}{c}{200 demos} &
      \multicolumn{3}{c}{500 demos} &
      \multicolumn{3}{c}{1000 demos} \\
    \cmidrule(lr){2-4}\cmidrule(lr){5-7}\cmidrule(lr){8-10}\cmidrule(lr){11-13}\cmidrule(lr){14-16}
    & MimicGen & Ours IL & Ours Ens
    & MimicGen & Ours IL & Ours Ens
    & MimicGen & Ours IL & Ours Ens
    & MimicGen & Ours IL & Ours Ens
    & MimicGen & Ours IL & Ours Ens \\
    \midrule
    Mug Cleanup
      & 21.2 $\pm$ 5.7 & 24.8 $\pm$ 6.9 & 33.6 $\pm$ 7.1
      & 37.2 $\pm$ 6.9 & 42.0 $\pm$ 6.1 & 38.0 $\pm$ 4.0
      & 50.0 $\pm$ 5.2 & 52.4 $\pm$ 5.3 & 42.8 $\pm$ 5.7
      & 61.6 $\pm$ 7.8 & 60.4 $\pm$ 10.1 & 46.0 $\pm$ 2.2
      & 65.6 $\pm$ 5.9 & 65.6 $\pm$ 3.2 & 58.0 $\pm$ 5.2 \\
    Kitchen
      & 92.0 $\pm$ 5.1 & 88.8 $\pm$ 9.0 & 56.4 $\pm$ 5.0
      & 98.4 $\pm$ 2.3 & 88.8 $\pm$ 1.6 & 60.0 $\pm$ 2.8
      & 99.6 $\pm$ 0.8 & 94.4 $\pm$ 4.1 & 62.8 $\pm$ 2.0
      & 98.8 $\pm$ 1.6 & 96.0 $\pm$ 4.2  & 63.2 $\pm$ 2.0
      & 100.0 $\pm$ 0.0 & 92.4 $\pm$ 4.1 & 63.6 $\pm$ 3.4 \\
    Coffee
      & 84.0 $\pm$ 5.2 & 86.4 $\pm$ 5.4 & 93.6 $\pm$ 2.3
      & 92.4 $\pm$ 3.2 & 89.6 $\pm$ 2.7 & 93.2 $\pm$ 3.2
      & 92.4 $\pm$ 4.5 & 96.0 $\pm$ 2.8 & 96.4 $\pm$ 0.8
      & 95.6 $\pm$ 2.9 & 95.2 $\pm$ 4.1 & 96.4 $\pm$ 2.0
      & 96.4 $\pm$ 2.3 & 99.2 $\pm$ 1.6 & 98.8 $\pm$ 1.0 \\
    \midrule\midrule
    \multirow{2}{*}{Task} &
      \multicolumn{3}{c}{25 demos} &
      \multicolumn{3}{c}{50 demos} &
      \multicolumn{3}{c}{100 demos} &
      \multicolumn{3}{c}{200 demos} &
      \multicolumn{3}{c}{400 demos} \\
    \cmidrule(lr){2-4}\cmidrule(lr){5-7}\cmidrule(lr){8-10}\cmidrule(lr){11-13}\cmidrule(lr){14-16}
    & ACT & Ours IL & Ours Ens
    & ACT & Ours IL & Ours Ens
    & ACT & Ours IL & Ours Ens
    & ACT & Ours IL & Ours Ens
    & ACT & Ours IL & Ours Ens \\
    \midrule
    Stack
      & 0.0 $\pm$ 0.0 & 0.8 $\pm$ 1.0 & 9.2 $\pm$ 4.7
      & 2.4 $\pm$ 2.0 & 4.8 $\pm$ 4.1 & 29.6 $\pm$ 7.8
      & 25.2 $\pm$ 4.1 & 34.8 $\pm$ 8.1 & 80.8 $\pm$ 4.7
      & 52.4 $\pm$ 14.1 & 61.2 $\pm$ 6.4 & 88.0 $\pm$ 4.9
      & 79.6 $\pm$ 5.0 & 92.0 $\pm$ 4.0 & 98.4 $\pm$ 1.5 \\
    Pick \& Place
      & 17.2 $\pm$ 7.1 & 30.4 $\pm$ 6.0 & 48.8 $\pm$ 5.6
      & 48.8 $\pm$ 10.5 & 56.0 $\pm$ 5.8 & 74.4 $\pm$ 10.4
      & 64.8 $\pm$ 4.7 & 79.6 $\pm$ 6.7 & 93.2 $\pm$ 3.0
      & 77.2 $\pm$ 3.2 & 92.0 $\pm$ 6.7 & 98.0 $\pm$ 1.3
      & 87.2 $\pm$ 4.1 & 98.4 $\pm$ 0.8 & 98.8 $\pm$ 1.0 \\
    \bottomrule
  \end{tabular}}
\end{table*}

\begin{table*}[t]
  \caption{Task success rates (\%) on additional tasks. Values are mean $\pm$  standard deviation}
  \label{tab:additional-tasks}
  \centering
  \small
  \setlength{\tabcolsep}{6pt}
  \renewcommand{\arraystretch}{1.1}

  \resizebox{\textwidth}{!}{%
  \begin{tabular}{l *{5}{cc}}
    \toprule
    \multirow{2}{*}{Task} &
      \multicolumn{2}{c}{25 demos} &
      \multicolumn{2}{c}{50 demos} &
      \multicolumn{2}{c}{100 demos} &
      \multicolumn{2}{c}{200 demos} &
      \multicolumn{2}{c}{400 demos} \\
    \cmidrule(lr){2-3}\cmidrule(lr){4-5}\cmidrule(lr){6-7}\cmidrule(lr){8-9}\cmidrule(lr){10-11}
    & IL & Ensembled
    & IL & Ensembled
    & IL & Ensembled
    & IL & Ensembled
    & IL & Ensembled \\
    \midrule
    Stack Flipped
      & 0.0 $\pm$ 0.0 & 4.4 $\pm$ 2.3
      & 2.8 $\pm$ 1.0 & 18.4 $\pm$ 4.5
      & 10.8 $\pm$ 2.0 & 36.0 $\pm$ 3.3
      & 18.4 $\pm$ 9.9 & 47.6 $\pm$ 2.9
      & 50.8 $\pm$ 3.9 & 54.8 $\pm$ 6.1 \\
    Stack Walking
      & 0.8 $\pm$ 1.0 & 3.2 $\pm$ 2.7
      & 5.2 $\pm$ 1.6 & 11.6 $\pm$ 4.8
      & 32.0 $\pm$ 7.8 & 34.4 $\pm$ 8.9
      & 43.6 $\pm$ 10.1 & 46.4 $\pm$ 13.0
      & 76.8 $\pm$ 5.3 & 59.6 $\pm$ 10.8 \\
    \bottomrule
  \end{tabular}}
\end{table*}

\subsubsection{Data Generation}
The simulation results show that our LLM-based data generation method significantly outperformed the MimicGen and OneACTPlay baselines, with the optimized LLM achieving a higher success rate on all tasks. Additionally, the total success rate, which includes the exploration phase of the optimization method, also beat the baseline in all tasks except for the Stack task. Importantly, our method achieved this improvement over the previous methods despite the prior methods using a human expert to manually annotate the trajectories. In contrast, our method required no human input other than a short (one sentence) description of the task. Despite this major limitation, our system outperformed the baselines, showing that through our optimization process, LLM Trainer can outperform human experts.

Our success rate results also illustrate the importance of our optimization method. For all tasks, we found that using the best annotation found using our optimization method led to a 2-3$\times$ increase in generation success rate.

Finally, the alternating stack task illustrates how the reasoning of the LLM allows our method to be more generalizable. Unlike the baseline methods, which rely on hardcoded object-centered transforms, our method is able to adapt based on purely visual cues. The LLM is able to reason that if the shaded green block appears on top of the shaded blue block, then the order in which the blocks are stacked must be flipped. Our system is then able to dynamically adjust its choice of waypoints accordingly - a level of flexibility that prior methods lacked.  

\subsubsection{Trained Model Performance}
Comparing the performance of imitation learning agents trained on data generated by our method to that of agents trained on data generated by the baseline methods, we see that our method performs roughly the same as MimicGen, and moderately outperforms OneACTPlay. These results show that our method generates high-quality data useful for training imitation learning policies. Somewhat unexpectedly, models trained using our data consistently outperformed identical models trained using data generated using OneACTPlay's method. We think this improvement is due to OneACTPlay's reliance on a single demonstration. Although our method also only uses a single human demonstration, the natural stochasticity of the LLM generation process increases the variation of the generated dataset, thereby making it more useful for training. We believe we didn't see a similar result for MimicGen due to that method's use of 10 human demonstrations, which adds additional variety to the baseline. 

\subsubsection{Ensembling Performance}
Examining the success rate of our ensembling method, we see that ensembling significantly improved the performance of IL agents in low data regimes, but occasionally hindered them in high data regimes where the trained IL agent was able to solve the task on its own. Interestingly, this decrease in performance did not apply to the block manipulation tasks. We believe this discrepancy stems from the fact that block manipulation failures are recoverable, while many failures in the MimicGen tasks are unrecoverable. For example, if the robot drops a block, it can easily pick it back up, while if the robot drops a mug and the mug lands on its side, the robot will be unable to retrieve it. Therefore, on the MimicGen tasks, the ensembled policy can be limited by the performance of the feedforward agent - once the ensembled agent realizes the feedforward agent has made a mistake, it may already be too late. 

Comparing the ensembled agent's performance to that of pure LLM feedforward control (using best annotation from the data generation process), we see that the ensembled agent often underperforms. This is in part due to the static nature of our evaluation tasks, which makes them well suited for feedforward agents. If the scene is dynamic, however, the lack of feedback control quickly becomes detrimental. To explore this, we evaluated the LLM feedforward agent on the walking stack task. In this environment, the LLM feedforward agent achieved a success rate of only 14\%, illustrating the brittleness of these feedforward policies. 

\subsection{Hardware Experiments}
\begin{figure}[thpb]
      \centering
      \includegraphics[width=\linewidth]{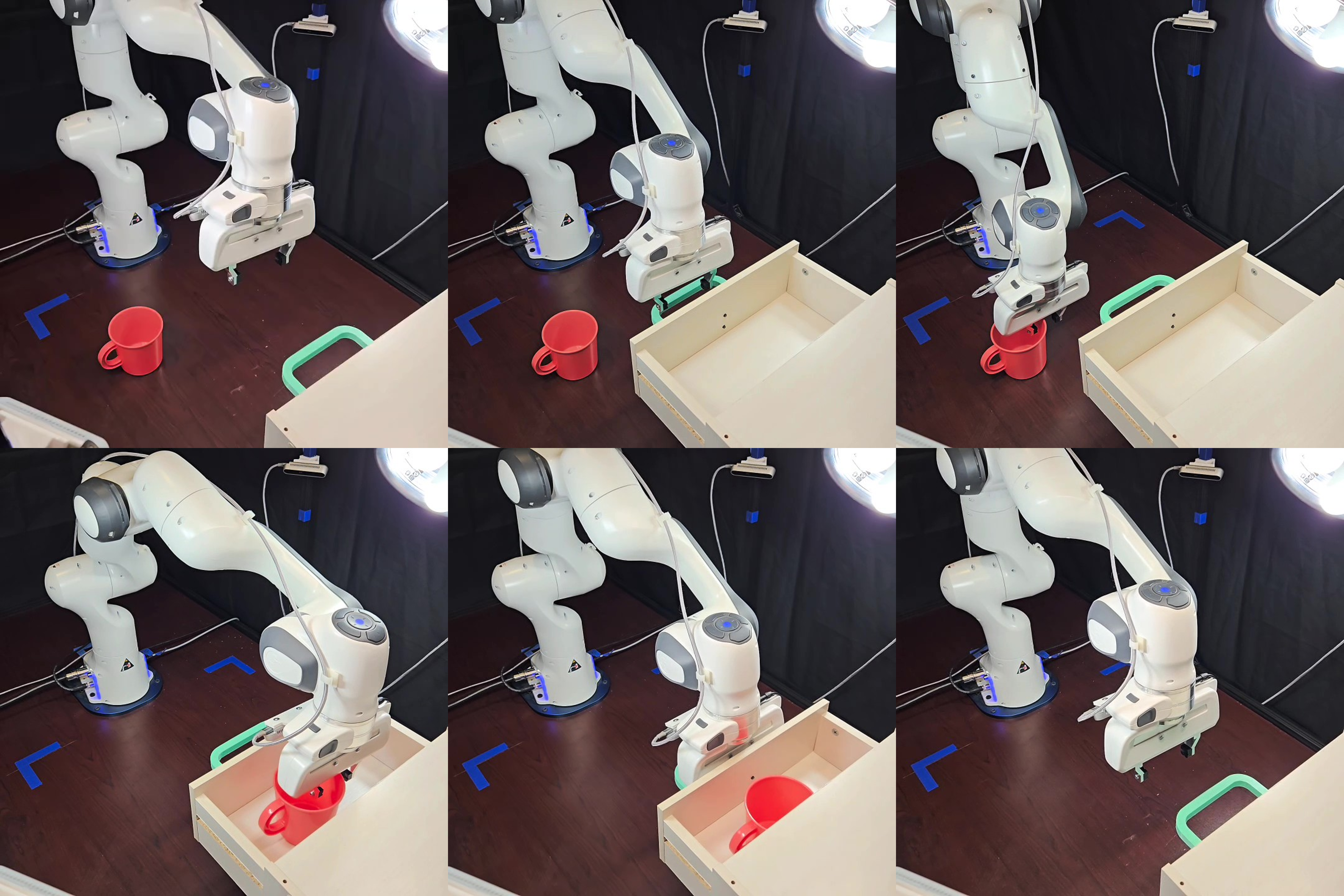}
      \caption{Mug Clean-up task used for the hardware experiments. For this task, the robot must open the drawer, pick up the mug, place it into the drawer, and then close the drawer.}
      \label{fig:hardware-task}
\end{figure}

To validate that our data generation method works on hardware, we evaluated it using a Franka Emika Panda robot, completing a mug-cleanup task. In this task, the robot must open a drawer, pick up the mug from the scene, place it in the drawer, and then close the drawer. The location of the mug in the scene was randomized (placed in a 20 cm $\times$ 30 cm region in front of the robot), as was the rotation of the mug ($\pm$ 45 deg). A diagram of this task can be seen in Fig. \ref{fig:hardware-task}. 

First, we used our method to generate 100 successful demonstrations, generating 32 failures along the way, resulting in a total success rate of 75.8\%. During this process, the best annotation (the annotation with the highest success rate) was used 95 times, generating 78 successful demonstrations - an average success rate of 82\%. We also did a separate experiment to evaluate the success rate of our method without optimization, and found a success rate of 45\%. We then trained an IL agent from \cite{george2023one} using the 100 successful demonstrations. We evaluated the IL agent, the feedforward LLM agent (using the best annotation), and the ensembled agent, running 20 trials for each. We found the IL agent had a success rate of 60\%, the LLM feedforward agent had a success rate of 80\%, and the ensembled agent had a success rate of 85\%.

These results align with our findings from simulation, with the optimization method significantly improving the data generation success rate, and the ensemble agent performing slightly better than the LLM feedforward agent. Interestingly, our data generation method worked better for our physical mug clean-up task than for the simulated version of this task. We think this is likely because the location of the mug in simulation is not the geometric center of the mug, but rather the location of the handle. In contrast, the location of the mug for the physical experiments was the centroid of the point cloud. We believe this more consistent location made understanding the mug's pose easier for the LLM.

\section{Conclusion}
We introduced \emph{LLM Trainer}, an autonomous demonstration generation system that leverages LLM world knowledge to adapt a small number of human demonstrations (as few as one) to new scenes, enabling automated collection of robot training data without manual annotation or task-specific hard-coding. In addition to removing the reliance on human input, the inclusion of LLMs enables the generation system to use world knowledge to dynamically adapt demonstrations based on context, providing our system with flexibility that human annotation based baselines lack. Using Thompson sampling to optimize reusable LLM annotations, our method significantly increases generation success rates, reaching success rates two to three times higher than those achieved using naïve annotations, and allowing our system to surpass expert engineered baselines. The optimized LLM feedforward policy is also a capable controller in its own right, and when ensembled with a feedback imitation learning agent, significantly boosts success rates in low-data regimes.

Crucially, we also show that the demonstrations produced by LLM Trainer are useful for robot learning. Imitation learning agents trained on the generated data achieve performance on par with baseline methods, confirming that our pipeline produces high-quality trajectories for policy learning. Finally, we validate end-to-end feasibility on hardware, demonstrating reliable data generation (75.8\% average generation success rate) on a long-horizon mug clean-up task using a Franka Emika Panda robot. Using this data, we were able to train an imitation learning agent to complete the task, and used this trained agent to validate our ensembling approach, achieving a success rate of 85\%.

\textbf{Limitations and future work.}
The main limitation of our work is its reliance on robot rollouts to validate proposed trajectories and record demonstrations. While straightforward in simulation, on hardware this process is time-consuming and may require human supervision. Automated resetting and pre-execution validation of the proposed trajectory could mitigate this concern, but we leave this task to future work. Another limitation of our work is that our system is optimized to reach a target number of successful demonstrations with the fewest rollouts, ignoring LLM prompting cost and the downstream utility of an optimized feedforward policy. Future work may seek to incorporate these factors into a more holistic optimization objective, tailored to specific use cases.

In summary, LLM Trainer provides a robust, task-agnostic, generalizable solution for demonstration generation, capable of producing high-quality data for robot learning from a single, unannotated human demonstration and a short description of the task.

\bibliographystyle{IEEEtran}
\bibliography{references}
\vspace{12pt}

\end{document}